\documentclass{article}

\usepackage{PRIMEarxiv}

\usepackage[utf8]{inputenc}
\usepackage[T1]{fontenc}
\usepackage{hyperref}
\usepackage{url}
\usepackage{amsmath,amssymb,amsthm,bm,mathtools}
\usepackage{booktabs,array,multirow}
\usepackage{amsfonts}
\usepackage{nicefrac}
\usepackage{microtype}
\usepackage{graphicx}
\usepackage{enumitem}
\usepackage{xcolor}
\usepackage{float}
\usepackage{caption}

\hypersetup{colorlinks=true,linkcolor=blue,citecolor=blue,urlcolor=blue}
\newcommand{\safeincludegraphics}[2][]{%
  \IfFileExists{#2}{\includegraphics[#1]{#2}}{%
  \fbox{\parbox[c][0.22\textheight][c]{0.86\linewidth}{\centering Figure placeholder\\\texttt{\detokenize{#2}}}}}}

\title{A Boltzmann-Machine-Enhanced Transformer for DNA Sequence Classification}

\author{
  Zhixuan Cao \\
  Tsinghua University \\
  \texttt{czx666666nnn@163.com} \\
  \And
  Yishu Xu \\
  University of California, Berkeley \\
  \texttt{kris-xu111shu@berkeley.edu} \\
  \And
  Xuang Wu \\
  Tsinghua University \\
  \texttt{wxa22@mails.tsinghua.edu.cn} \\
}

\date{}

\begin{document}
\maketitle

\begin{abstract}
DNA sequence classification requires not only strong predictive accuracy but also the ability to reveal latent site interactions, combinatorial regulation, and epistasis-like higher-order dependencies. Although the standard Transformer provides powerful global modeling capacity, its softmax attention is continuous, dense, and only weakly constrained; it is therefore better interpreted as an information routing mechanism than as an explicit structure discovery process. In this paper, we propose a Boltzmann-machine-enhanced Transformer for DNA sequence classification. Building on standard multi-head attention, we introduce structured binary gating variables to represent latent query--key connections and impose priors over these connections through a Boltzmann-style energy function. Specifically, query--key similarity defines local bias terms, learnable pairwise interactions capture synergy and competition between edges, and explicit latent hidden units model higher-order combinatorial dependencies. Because exact posterior inference over discrete gating graphs is intractable, we adopt mean-field variational inference to estimate the approximate activation probability of each edge, and then combine it with Gumbel-Softmax to gradually compress continuous probabilities into near-discrete gates while preserving end-to-end differentiability. During training, we jointly optimize a classification loss and an energy loss so that the model not only predicts accurately but also prefers lower-energy, more stable, and more interpretable structures. We further provide a complete derivation from the energy function and variational free energy to mean-field fixed-point equations, Gumbel-Softmax relaxation, and the final joint objective. This framework offers a unified view of integrating Boltzmann machines, differentiable discrete optimization, and Transformers, and suggests a new direction for structured deep learning on biological sequences.
\end{abstract}


\section{Introduction}
DNA sequences contain abundant local motifs, long-range interactions, and higher-order combinatorial regulatory patterns. These properties jointly determine the difficulty of tasks such as regulatory element identification, transcription factor binding prediction, and sequence function classification. In recent years, deep learning has become the dominant paradigm for such problems, especially because it exhibits strong representation-learning ability on large-scale genomic sequences \cite{ji2021dnabert,kelley2021enformer,gresova2023genomic}. For tasks such as enhancer recognition, functional fragment classification, and transcription factor binding prediction, a useful model must provide not only strong discriminative performance but also some degree of structural interpretability, for example by revealing which sites exhibit synergistic behavior and which local patterns jointly influence the output label in a modular manner.

In recent years, the Transformer has become a foundational architecture for sequence modeling. Its self-attention mechanism can capture dependencies among sites at a global scale and has achieved remarkable progress in both natural language processing and biological sequence modeling \cite{vaswani2017attention,ji2021dnabert,kelley2021enformer}. However, in the standard Transformer, attention weights are generated through softmax normalization and therefore form a continuous, dense allocation of weights. Although such a mechanism is highly effective for prediction, it usually lacks explicit structural constraints and is consequently difficult to interpret directly in terms of whether a specific edge exists or which sites form a stable interaction graph.

From a biological perspective, one often hopes that the model will learn not merely a classifier but also a latent interaction graph. The edges in this graph should be statistically meaningful and ideally exhibit sparsity, synergy, competition, or modularity. To this end, we propose a Boltzmann-machine-enhanced Transformer that explicitly models query--key connections as gating variables and uses an energy function to characterize the plausibility of the entire gating graph. This modeling strategy is inspired by Boltzmann machines and energy-based models: the model learns not only the predictor itself but also a prior distribution that favors low-energy, low-conflict, and interpretable structures \cite{ackley1985boltzmann,hinton2010practical}.

The core idea of this paper is to replace standard softmax attention with a Boltzmann-style structural distribution. Rather than directly using softmax to normalize all keys associated with a query into a continuous weight distribution, we model whether a connection exists as a latent binary random variable and assign a Boltzmann distribution over the entire connection graph. In this way, attention becomes not only an information routing mechanism but also a process of structural discovery.

This formulation also introduces new challenges. First, the gating variables are discrete, so direct optimization would make gradients non-differentiable. Second, the gating variables are not independent; they are coupled through pairwise interactions and latent hidden units, which makes exact posterior inference generally intractable. To address these issues, we introduce two key techniques. First, we use mean-field variational inference to approximately solve the coupled system, converting the complex posterior into a fixed-point problem over continuous probabilities \cite{peterson1987mean,tanaka1998mean}. Second, we use Gumbel-Softmax to gradually push the soft structure toward a discrete $0/1$ configuration while preserving gradient flow and improving the sharpness and sparsity of the gating graph \cite{jang2017gumbel,maddison2017concrete}.

Moreover, classification loss alone cannot guarantee that the intermediate structure is biologically interpretable. We therefore introduce an energy loss in addition to the task loss, requiring the structure obtained from the current forward inference to have lower energy than negative structures generated by perturbation or sampling. As a result, the training objective is no longer merely to make the prediction correct, but to make the prediction correct through a plausible structure.

The main contributions of this paper are as follows:
\begin{enumerate}[label=(\arabic*)]
    \item We propose a Boltzmann-machine-enhanced Transformer for DNA sequence classification, extending standard softmax attention into structured gated attention equipped with an energy-based prior;
    \item We provide a complete derivation from the energy model to mean-field fixed-point iterations, clarifying the theoretical origin of the gating-probability updates;
    \item We combine Gumbel-Softmax with mean-field inference so that complex coupled near-discrete structures can be trained end to end;
    \item By jointly optimizing task loss and energy loss, we establish a unified optimization framework that balances predictive performance and structural plausibility;
    \item We discuss the relationship between the proposed method and the standard Transformer, and explain why structured attention is better suited to interpretable modeling and higher-order interaction learning.
\end{enumerate}

\section{Background and Related Work}
Recent work on DNA sequence modeling, interpretable attention, and alternatives to softmax has gradually converged into three technical trajectories: first, Transformer-based and foundation-model approaches for genomic sequence understanding; second, theoretical approaches that interpret attention as an energy model or a Boltzmann-like operator; and third, mechanism-design approaches that directly restructure, sparsify, or replace softmax attention. The method proposed here lies at the intersection of these three lines of research.

\subsection{DNA Sequence Modeling}
In biological sequence tasks, the Transformer has evolved from a generic sequence-modeling tool into an important class of genomic foundation models. DNABERT was the first to systematically transfer BERT-style pretraining to DNA sequences, emphasizing the importance of global context for recognizing promoters, splice sites, and transcription factor binding sites \cite{ji2021dnabert}. Enformer subsequently showed that long-range interactions are crucial for gene expression prediction, with an architecture capable of integrating sequence dependencies over distances of up to 100 kb, thereby demonstrating that local motifs alone are insufficient to explain complex regulatory relationships \cite{kelley2021enformer}. Building on this line of work, DNABERT-2 and Nucleotide Transformer extended the paradigm toward larger scale, multi-species learning, and multi-task transfer, highlighting the strengths of DNA foundation models in low-label settings, cross-task generalization, and variant-effect prediction \cite{zhou2023dnabert2,dallatorre2025nucleotide}. A common feature of this trajectory is that models have become increasingly adept at representing sequences, while their attention weights are still usually treated as continuous importance scores rather than explicit interpretable structural variables.

\subsection{The Connection Between Softmax and Boltzmann Formulations}
Another important line of work arises from theoretical reinterpretations of the attention mechanism itself. In their analysis of the expressive power of single-layer self-attention, Kajitsuka and Sato showed that softmax-weighted aggregation can be written in the form of a Boltzmann operator, and used this connection to establish stronger expressive results than earlier analyses based on hardmax approximations \cite{kajitsuka2024transformers}. Importantly, such work explains why standard softmax attention is effective; it does not propose a new Boltzmann-machine-style attention module. Closely related, Ota and Karakida studied a family of Boltzmann machines associated with attention through the lens of modern Hopfield networks, and proposed the attentional Boltzmann machine (AttnBM), thereby showing that attention modules can be embedded in a broader family of energy functions \cite{ota2023attention}. These results provide theoretical justification for rewriting or extending attention in Boltzmann-style structural terms, although they do not directly yield a softmax replacement tailored to DNA sequence classification.

\subsection{Softmax Alternatives and Structured Attention}
In addition to theoretical explanations of softmax, a substantial body of work has attempted to replace or modify softmax attention directly. Early methods focused primarily on computational efficiency and sparsity, for example by constructing linear attention through kernel tricks or reweighting mechanisms; among them, cosFormer redesigned the attention transformation to retain non-negativity while avoiding the quadratic complexity of standard softmax \cite{qin2022cosformer}. Subsequently, some studies challenged softmax more directly, such as replacing it with non-normalized activations like ReLU in vision Transformers and demonstrating feasibility on specific visual tasks \cite{shen2024reluattn}. More recently, BoltzFormer introduced Boltzmann distributions and annealing strategies into the sparse sampling process of Transformers for small-object image analysis, indicating that Boltzmann-inspired attention has begun to emerge at the level of concrete mechanism design \cite{zhao2025boltzmann}. However, these methods are mostly aimed at efficiency optimization, region sampling, or visual applications; they typically do not model query--key edges themselves as coupled binary random variables, nor do they explicitly introduce pairwise interactions and latent hidden units to capture higher-order structural dependence.

\subsection{Distinctions Between the Proposed Method and Prior Work}
Taken together, existing DNA foundation models demonstrate the effectiveness of Transformers for sequence discrimination and representation learning \cite{ji2021dnabert,kelley2021enformer,zhou2023dnabert2,dallatorre2025nucleotide}; energy-based and Hopfield/Boltzmann perspectives reveal theoretical connections between attention and Boltzmann structures \cite{kajitsuka2024transformers,ota2023attention}; and research on non-softmax attention shows that softmax is not the only viable normalization mechanism for attention \cite{qin2022cosformer,shen2024reluattn,zhao2025boltzmann}. The present work differs from these studies in that it is explicitly designed for DNA sequence classification and replaces standard softmax attention with a Boltzmann-inspired structured gating mechanism. This mechanism does not merely alter the normalization function; instead, it treats the query--key connectivity graph as a discrete structural object endowed with an energy-based prior, and achieves end-to-end training through mean-field inference and Gumbel-Softmax relaxation. In other words, this paper advances the view of attention from weight allocation to structural discovery, thereby preserving the Transformer's global modeling capacity while better aligning with the interpretive demands of latent interaction graphs in biological sequences.

\section{Method}
\subsection{Task Definition and Overall Architecture}
Given a DNA sequence $x = (x_1, x_2, \dots, x_L)$ of length $L$, where each position takes a value from the nucleotide alphabet $\{A, C, G, T, N\}$, the goal is to predict a binary label $y \in \{0,1\}$. The model first maps the input sequence to discrete embeddings, then extracts local motif features through a lightweight one-dimensional convolutional module, subsequently performs global modeling with a Transformer encoder, and finally outputs the logit through masked mean pooling and an MLP classification head.

The principal difference between the proposed model and a standard Transformer lies in the attention layer. In a standard Transformer, attention is generated directly by softmax; in contrast, we introduce within each head a latent binary gating graph to represent whether a connection is established between a query and a key. The entire gating graph is modeled by a Boltzmann-style energy function and is inferred through a trainable approximation based on mean-field inference and Gumbel-Softmax.

\subsection{From Softmax Attention to Structured Gated Attention}
\subsubsection{Standard Softmax Attention}
For the $h$-th head, let $\mathbf{q}_{ht} \in \mathbb{R}^{d_h}$ denote the $t$-th query, $\mathbf{k}_{hs} \in \mathbb{R}^{d_h}$ denote the $s$-th key, and $\mathbf{v}_{hs} \in \mathbb{R}^{d_h}$ denote the corresponding value. Standard softmax attention is defined as
\begin{equation}
A_{hts} = \frac{\exp\!\left(\mathbf{q}_{ht}^{\top} \mathbf{k}_{hs} / \sqrt{d_h}\right)}{\sum_{s'} \exp\!\left(\mathbf{q}_{ht}^{\top} \mathbf{k}_{hs'} / \sqrt{d_h}\right)}.
\end{equation}
which yields the head output
\begin{equation}
\mathbf{o}_{ht} = \sum_s A_{hts} \, \mathbf{v}_{hs}.
\end{equation}
A key characteristic of this mechanism is that all weights within a query row are normalized onto a probability simplex, so the operation is better viewed as weight allocation rather than structural decision-making. Moreover, aside from competition induced by normalization, there is no explicit coupling among different edges.

\subsubsection{Gating Graph Variables}
To endow attention with a clearer structural interpretation, we introduce binary gating variables
\begin{equation}
z_{hts} \in \{0,1\},
\end{equation}
where $z_{hts}=1$ indicates that the connection between query position $t$ and key position $s$ is activated in head $h$, and $z_{hts}=0$ indicates that the connection is inactive. We therefore wish the attention output to be controlled by the gating graph:
\begin{equation}
\mathbf{o}_{ht} = \frac{\sum_s z_{hts} \, \mathbf{v}_{hs}}{\sum_s z_{hts} + \varepsilon}.
\end{equation}
Because direct discrete optimization over $z$ is infeasible, we introduce a continuous variable $s_{hts}$ to approximate the marginal posterior probability of $z_{hts}$:
\begin{equation}
s_{hts} \approx q(z_{hts}=1).
\end{equation}
In what follows, $s$ serves as the soft version of the structured attention graph.

\subsection{Boltzmann-Style Structural Distribution}
\subsubsection{Definition of the Structural Distribution}
We define the probability of the entire gating graph $z$ using a Boltzmann distribution, consistent with the classical energy-based formulation of Boltzmann machines \cite{ackley1985boltzmann,hinton2010practical}:
\begin{equation}
p(z\mid x) = \frac{1}{Z(x)} \exp\!\bigl(-E(z;x)\bigr),
\end{equation}
where $E(z;x)$ is the energy function conditioned on input $x$, and $Z(x)$ is the partition function:
\begin{equation}
Z(x) = \sum_z \exp\!\bigl(-E(z;x)\bigr).
\end{equation}
Because $z$ is a high-dimensional binary tensor, the number of possible structures grows exponentially, making exact computation of $Z(x)$ and the marginal probabilities generally intractable.

\subsubsection{Local Bias Term}
We first define a local bias term based on query--key similarity:
\begin{equation}
h_{hts} = \frac{\mathbf{q}_{ht}^{\top} \mathbf{k}_{hs}}{\sqrt{d_h}}.
\end{equation}
This quantity describes the tendency of an edge to be activated on the basis of local similarity alone, without considering other edges. The corresponding bias-energy term is defined as
\begin{equation}
E_{\mathrm{bias}}(z;x) = - \sum_h \sum_t \sum_s h_{hts} z_{hts}.
\end{equation}
Lower energy indicates that the structure more strongly favors activating edges with high similarity.

\subsubsection{Pairwise Interaction Term}
To explicitly model coupling between edges, we introduce pairwise interactions within each query row. Unlike traditional Boltzmann machines, which directly learn free coupling coefficients $W_{ij}$, we express the coupling strength here as a dynamic quantity induced by the attention representations. For the $h$-th head and the $t$-th query, we define the coupling between candidate key positions $s$ and $s'$ as
\begin{equation}
J_{ss'}^{(h,t)} = \mathbf{k}_{hs}^{\top} \mathbf{W}_{\mathrm{sym}}^{(h,t)} \mathbf{k}_{hs'},
\end{equation}
where $\mathbf{W}_{\mathrm{sym}}^{(h,t)}$ is a symmetric parameter matrix. In implementation, one may use diagonal parameterization or explicit symmetrization to ensure that
\begin{equation}
J_{ss'}^{(h,t)} = J_{s's}^{(h,t)},
\end{equation}
thereby maintaining consistency with the symmetric-coupling assumption in the energy function. This construction means that $J_{ss'}^{(h,t)}$ is no longer a static, input-independent parameter; instead, it is determined by the compatibility between the two key representations in the current sample, reflecting whether they tend to co-occur under the present attention context.

On this basis, the pairwise energy term is written as
\begin{equation}
E_{\mathrm{pair}}(z;x) = -\frac{1}{2} \sum_h \sum_t \sum_{s \neq s'} J_{ss'}^{(h,t)} z_{hts} z_{hts'}.
\end{equation}
Here, $z_{hts}$ indicates whether query position $t$ connects to key position $s$. When $J_{ss'}^{(h,t)}>0$, the two edges tend to be activated synergistically; when $J_{ss'}^{(h,t)}<0$, they tend to be mutually exclusive. Accordingly, this term can be interpreted as replacing the conventional pairwise interaction parameters in a Boltzmann machine with coupling terms induced by key--key interactions. It explicitly models cooperation and competition among candidate connections. This is also a key distinction from standard softmax attention: the model no longer decides solely on the basis of local similarity for individual edges, but instead directly encodes the structural prior that edges can influence one another into the energy function.

\subsubsection{Latent Hidden Units Term}
To model higher-order combinatorial relationships, we further introduce explicit latent variables $u_{htm}\in\{0,1\}$, where $m=1,\dots,M$ indexes the $m$-th latent hidden unit. We define
\begin{equation}
E_{\mathrm{latent}}(z,u;x) = - \sum_h \sum_t \sum_m b_{htm}^{(u)} u_{htm} - \sum_h \sum_t \sum_s \sum_m W_{sm}^{(h,t)} z_{hts} u_{htm}.
\end{equation}
This term means that when a group of edges jointly supports the activation of a latent unit, that latent unit in turn reinforces a related set of edges, thereby expressing modular or combinatorial regulatory patterns. After marginalizing out $u$, the latent units induce effective higher-order interactions over $z$.

\subsubsection{Overall Energy Function}
The overall energy is defined as
\begin{equation}
E(z,u;x) = E_{\mathrm{bias}}(z;x) + E_{\mathrm{pair}}(z;x) + E_{\mathrm{latent}}(z,u;x).
\end{equation}
Marginalizing out $u$ yields an effective energy function over $z$. Because exact marginalization remains difficult, we adopt a mean-field approximation over both $z$ and $u$ during inference.

\subsection{Mean-Field Variational Inference}
\subsubsection{Approximate Distribution}
We adopt a factorized approximation (naive mean-field approximation) \cite{peterson1987mean,tanaka1998mean}:
\begin{equation}
q(z,u) = \prod_h \prod_t \prod_s q_{hts}(z_{hts}) \cdot \prod_h \prod_t \prod_m r_{htm}(u_{htm}),
\end{equation}
where
\begin{equation}
q_{hts}(z_{hts}=1)=s_{hts}, \qquad r_{htm}(u_{htm}=1)=r_{htm}.
\end{equation}
Here, $s_{hts}$ and $r_{htm}$ are the mean-field parameters for the gated edges and latent units, respectively.

\subsubsection{Variational Free Energy}
Obtaining the optimal approximation by minimizing $\mathrm{KL}(q\|p)$ is equivalent to minimizing the variational free energy:
\begin{equation}
\mathcal{F}(q)=\mathbb{E}_q\bigl[E(z,u;x)\bigr]-H(q),
\end{equation}
where $H(q)$ is the entropy term. Because $q$ is a factorized Bernoulli distribution, its entropy is given by
\begin{align}
H(q) = {} & - \sum_h \sum_t \sum_s \Bigl[s_{hts}\log s_{hts} + (1-s_{hts})\log(1-s_{hts})\Bigr] \\
& - \sum_h \sum_t \sum_m \Bigl[r_{htm}\log r_{htm} + (1-r_{htm})\log(1-r_{htm})\Bigr].
\end{align}

\subsubsection{Expected Energy}
Taking the expectation of the bias term gives
\begin{equation}
\mathbb{E}_q\bigl[E_{\mathrm{bias}}\bigr] = - \sum_h \sum_t \sum_s h_{hts} s_{hts}.
\end{equation}
For the pairwise term, under the mean-field independence approximation, we have
\begin{equation}
\mathbb{E}_q[z_{hts} z_{hts'}] \approx s_{hts} s_{hts'},
\end{equation}
therefore
\begin{equation}
\mathbb{E}_q\bigl[E_{\mathrm{pair}}\bigr] = -\frac{1}{2} \sum_h \sum_t \sum_{s \neq s'} J_{ss'}^{(h,t)} s_{hts} s_{hts'}.
\end{equation}
For the latent term, we have
\begin{equation}
\mathbb{E}_q\bigl[E_{\mathrm{latent}}\bigr] = - \sum_h \sum_t \sum_m b_{htm}^{(u)} r_{htm} - \sum_h \sum_t \sum_s \sum_m W_{sm}^{(h,t)} s_{hts} r_{htm}.
\end{equation}
Summing these three components yields the total expected energy.

\subsubsection{Derivation of the Fixed-Point Update for $s$}
Taking the partial derivative of $\mathcal{F}(q)$ with respect to a particular $s_{hts}$ gives
\begin{equation}
\frac{\partial \mathcal{F}}{\partial s_{hts}} = - h_{hts} - \sum_{s' \neq s} J_{ss'}^{(h,t)} s_{hts'} - \sum_m W_{sm}^{(h,t)} r_{htm} + \log\frac{s_{hts}}{1-s_{hts}}.
\end{equation}
Setting it to zero yields
\begin{equation}
\log\frac{s_{hts}}{1-s_{hts}} = h_{hts} + \sum_{s' \neq s} J_{ss'}^{(h,t)} s_{hts'} + \sum_m W_{sm}^{(h,t)} r_{htm},
\end{equation}
that is,
\begin{equation}
s_{hts} = \sigma\!\left(h_{hts} + \sum_{s' \neq s} J_{ss'}^{(h,t)} s_{hts'} + \sum_m W_{sm}^{(h,t)} r_{htm}\right),
\end{equation}
where $\sigma(\cdot)$ denotes the sigmoid function. This is the core mean-field update equation used in the present work.

\subsubsection{Derivation of the Fixed-Point Update for $r$}
Similarly, taking the derivative of $\mathcal{F}(q)$ with respect to $r_{htm}$ and setting it to zero gives
\begin{equation}
\log\frac{r_{htm}}{1-r_{htm}} = b_{htm}^{(u)} + \sum_s W_{sm}^{(h,t)} s_{hts},
\end{equation}
thus
\begin{equation}
r_{htm} = \sigma\!\left(b_{htm}^{(u)} + \sum_s W_{sm}^{(h,t)} s_{hts}\right).
\end{equation}
Accordingly, the mean-field parameters for $z$ and $u$ form a coupled fixed-point system.

\subsubsection{Iterative Solution}
Given the initialization
\begin{equation}
s^{(0)} = \sigma(h),
\end{equation}
we then update alternately according to
\begin{align}
r^{(k)} &= \sigma\!\bigl(b^{(u)} + W^{\top} s^{(k)}\bigr), \\
s^{(k+1)} &= \sigma\!\bigl(h + J s^{(k)} + W r^{(k)}\bigr).
\end{align}
After $K$ iterations, we obtain a self-consistent soft structural solution $s^{(K)}$. This is the structural probability graph used in the current forward pass, namely the positive-phase structure $s_{\mathrm{pos}}$.

\subsection{Replacing Softmax with Boltzmann-Machine Gating}
Once $s$ is obtained, we no longer use softmax weights; instead, we directly use the structural probability graph to control information aggregation:
\begin{equation}
\mathbf{o}_{ht} = \frac{\sum_s s_{hts} \, \mathbf{v}_{hs}}{\sum_s s_{hts} + \varepsilon}.
\end{equation}
This step is important for three reasons:
\begin{enumerate}[label=(\arabic*)]
    \item whether each edge is activated is determined by the energy model and coupled inference, rather than solely by local $\mathbf{q}\cdot\mathbf{k}$ similarity;
    \item multiple edges within the same row can be strongly activated simultaneously, without being constrained by the softmax simplex;
    \item the resulting graph structure can exhibit sparsity, modularity, and interpretability.
\end{enumerate}
Thus, the proposed method can be regarded as a structured extension that replaces standard softmax attention with Boltzmann-machine-gated attention.

\subsection{Approximate Discretization via Gumbel-Softmax}
\subsubsection{Construction of Binary Classification Logits}
Although $s$ provides a reasonable posterior approximation, it remains continuous. To make the structure closer to an explicitly discrete graph, we construct binary classification logits:
\begin{equation}
\alpha_{hts}^{(0)} = \log(1-s_{hts}), \qquad \alpha_{hts}^{(1)} = \log(s_{hts}).
\end{equation}

\subsubsection{Gumbel-Softmax Sampling}
Let $g_{hts}^{(0)}$ and $g_{hts}^{(1)}$ be independent Gumbel noises. Then
\begin{equation}
\tilde{z}_{hts}^{(c)} = \frac{\exp\!\left((\alpha_{hts}^{(c)} + g_{hts}^{(c)})/\tau\right)}{\sum_{c'\in\{0,1\}} \exp\!\left((\alpha_{hts}^{(c')} + g_{hts}^{(c')})/\tau\right)}.
\end{equation}
We take the component with $c=1$ as the approximate gate:
\begin{equation}
\tilde{z}_{hts} = \tilde{z}_{hts}^{(1)}.
\end{equation}
When the temperature $\tau$ is high, the output is relatively smooth; as $\tau$ gradually decreases, the output approaches a one-hot vector. In later stages of training, a straight-through estimator can be employed so that the forward pass uses hard binary gates while the backward pass continues to propagate gradients through the soft relaxation.\\
The gradient provided by Gumbel-Softmax is not an unbiased gradient of the original discrete random variable, but it is a low-variance and stable differentiable approximation \cite{jang2017gumbel,maddison2017concrete}. For discriminative models such as ours, in which the intermediate structure serves as a latent variable, this biased but stable gradient is often more suitable for training than high-variance estimators of exact discrete gradients.

\subsection{Positive and Negative Structures for Energy Learning}
\subsubsection{Positive-Phase Structure}
The positive-phase structure $s_{\mathrm{pos}}$ is obtained from the current model through mean-field fixed-point iterations and can be further compressed into a near-discrete version by Gumbel-Softmax. It represents the structure that the model currently considers most plausible under the present parameters.

\subsubsection{Negative-Phase Structure}
To learn the structural energy, we also require a negative-phase structure $s_{\mathrm{neg}}$. We consider two approaches:
\begin{enumerate}[label=(\arabic*)]
    \item local random perturbation: randomly flipping a subset of edges in the neighborhood of $s_{\mathrm{pos}}$ to obtain a locally corrupted structure;
    \item an external sampler: optionally using approximate solvers such as simulated annealing to search for more challenging negative samples within local Ising subproblems; this strategy of enhancing structural search with external optimizers also resonates with recent work on quantum-boosted high-fidelity deep learning frameworks \cite{Wang2025QuantumBoostedHD}.
\end{enumerate}
Although random perturbation is simple, it offers the advantages of low variance and stable training, and can be understood as a form of local contrastive learning in the neighborhood of the current structure.

\subsection{Joint Loss Function}
\subsubsection{Task Loss}
For the final binary classification logit $\hat{y}$, we use binary cross-entropy:
\begin{equation}
\mathcal{L}_{\mathrm{task}} = -\Bigl[y\log\sigma(\hat{y}) + (1-y)\log\bigl(1-\sigma(\hat{y})\bigr)\Bigr].
\end{equation}

\subsubsection{Energy Loss}
Let the energy of the positive-phase structure be $E(s_{\mathrm{pos}})$ and that of the negative-phase structure be $E(s_{\mathrm{neg}})$. We then define a margin-ranking-style energy loss:
\begin{equation}
\mathcal{L}_{\mathrm{energy}} = \max\bigl(0, E(s_{\mathrm{pos}}) - E(s_{\mathrm{neg}}) + m\bigr),
\end{equation}
where $m>0$ is the margin. This loss encourages
\begin{equation}
E(s_{\mathrm{pos}}) + m \le E(s_{\mathrm{neg}}).
\end{equation}
That is, the structure obtained in the current forward pass should have lower energy than the negative-phase structure.

\subsubsection{Overall Loss}
The total loss is written as
\begin{equation}
\mathcal{L} = \mathcal{L}_{\mathrm{task}} + \lambda \, \mathcal{L}_{\mathrm{energy}}.
\end{equation}
Its gradient with respect to the parameters $\theta$ is
\begin{equation}
\nabla_{\theta} \mathcal{L} = \nabla_{\theta} \mathcal{L}_{\mathrm{task}} + \lambda \, \nabla_{\theta} \mathcal{L}_{\mathrm{energy}}.
\end{equation}
Thus, parameter updates are jointly driven by the task objective and the structural objective.

\subsubsection{Optimization Perspective}
This objective is equivalent to a regularized joint optimization problem: among all parameter settings that yield low task loss, the model prefers those associated with lower structural energy. In other words, $\mathcal{L}_{\mathrm{task}}$ enforces predictive correctness, whereas $\mathcal{L}_{\mathrm{energy}}$ enforces structural correctness.

\subsubsection{Probabilistic Interpretation}
If the energy is viewed as the negative log-probability of a structural prior, namely
\begin{equation}
E(s) \approx -\log p(s),
\end{equation}
then the overall objective can be approximately written as
\begin{equation}
\mathcal{L} \approx -\log p(y\mid x,s) - \lambda \log p(s).
\end{equation}
This shows that the model is effectively jointly optimizing two probabilistic objectives: predictive correctness and structural plausibility.

\begin{figure}[!htbp]
\centering
\safeincludegraphics[width=0.92\textwidth]{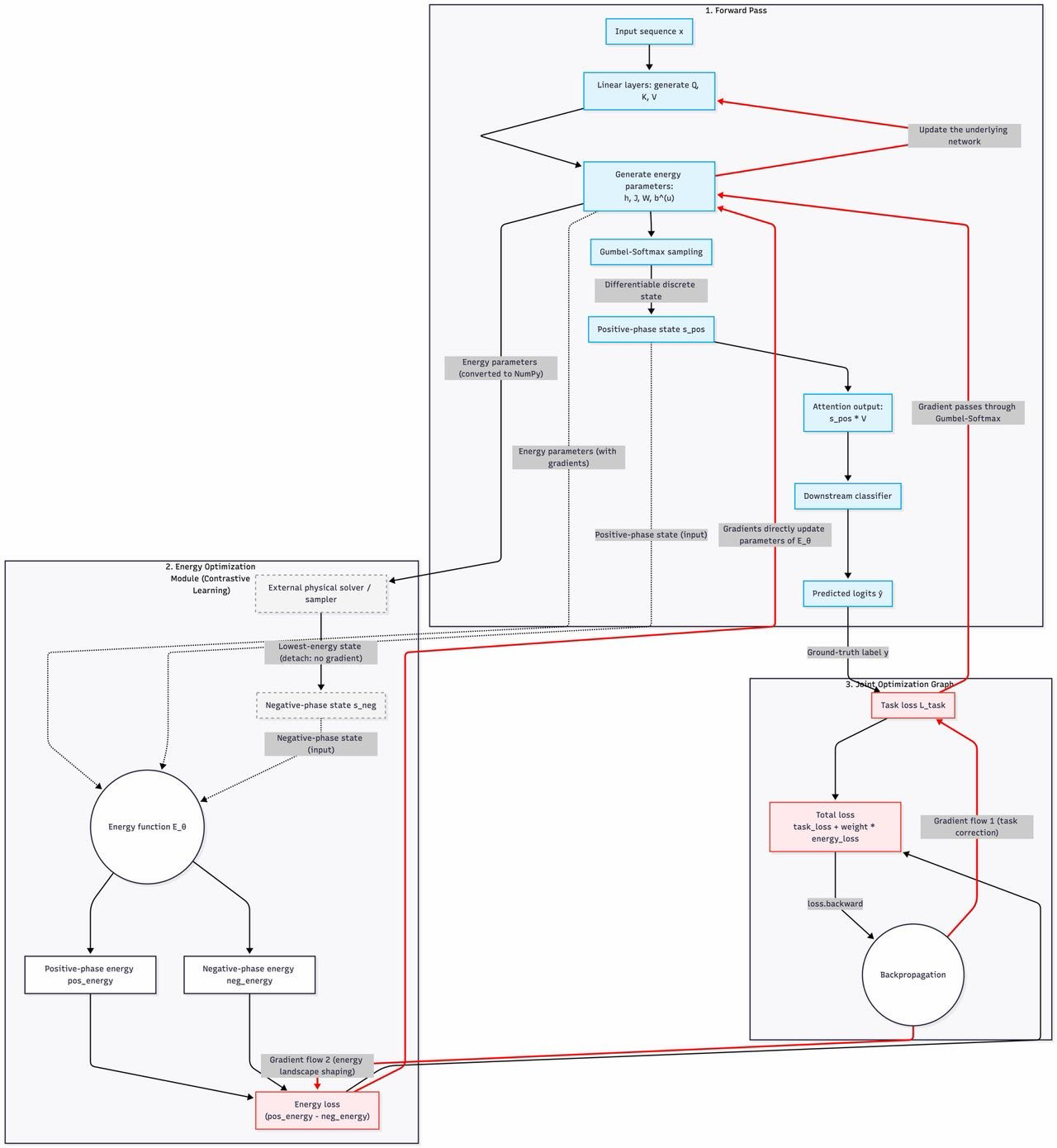}
\caption{End-to-end training pipeline of the proposed BM-Transformer. The forward branch generates $Q$, $K$, and $V$, constructs a differentiable positive-phase structure through Gumbel-Softmax, and feeds the resulting gated representation into the downstream classifier. In parallel, the energy-learning branch evaluates positive- and negative-phase structures with the energy function, where the negative phase can optionally be obtained from an external solver or sampler. The final objective jointly combines task loss and energy loss, and gradients flow back through both branches to update the underlying network and the energy parameters.}
\label{fig:bm_training_pipeline}
\end{figure}

Figure~\ref{fig:bm_training_pipeline} presents the overall computational graph of the proposed model during training. It simultaneously illustrates the forward classification branch, the energy-learning branch associated with positive- and negative-phase structures, and the joint backpropagation path induced by the task loss and energy loss, thereby clarifying the roles of Gumbel-Softmax, the energy function, and optional external solvers within a unified training framework.

\subsection{Training Strategy}
To avoid instability in the early stage of training, we adopt a curriculum-style strategy that proceeds from soft to hard structures:
\begin{enumerate}[label=(\arabic*)]
    \item in the initial stage, only the soft mean-field structure is used, without imposing a strong energy constraint;
    \item the Gumbel temperature $\tau$ is gradually decreased during training;
    \item after the model acquires basic discriminative ability, the weight $\lambda$ of the energy loss is gradually increased;
    \item in the later stage of training, hard sampling is enabled so that the forward gates more closely approximate an explicitly discrete graph structure.
\end{enumerate}
This strategy allows the model to first learn to solve the task correctly and then learn to do so through a clearer structure.

\section{Relationship to the Standard Transformer}
The proposed model is not intended to replace the Transformer, but rather to provide a structured extension of its attention mechanism. If structured gating, mean-field inference, Gumbel-Softmax, and the energy loss are all disabled, the model naturally reduces to a standard MultiheadAttention baseline.

The standard Transformer is already a very strong baseline for many DNA classification tasks because the convolutional front end can extract local motifs, self-attention is sufficient to model medium- and long-range dependencies, and the optimization objective is simpler and more stable. By contrast, the added value of the proposed method lies primarily in the following aspects:
\begin{enumerate}[label=(\arabic*)]
    \item transforming continuous and weakly constrained soft attention into a gated graph equipped with a structural prior;
    \item explicitly representing synergy and competition among edges;
    \item expressing higher-order combinatorial regulation through latent hidden units;
    \item improving the sparsity and interpretability of the intermediate structure.
\end{enumerate}
Accordingly, the proposed method is more suitable for settings in which one seeks both discriminative performance and structural interpretability.

\section{Experimental Design}
\subsection{Dataset and Task}
We evaluate the proposed model on the \texttt{Genomic\_Benchmarks\_human\_enhancers\_cohn} dataset for binary DNA sequence classification. This dataset is drawn from the Genomic Benchmarks collection \cite{gresova2023genomic} and contains a DNA sequence field \texttt{seq} together with a corresponding binary label field \texttt{label}. The training set contains 20,843 samples and the test set contains 6,948 samples. Given an input sequence $x$, the model must determine whether it belongs to the target functional category; accordingly, we treat the task uniformly as supervised binary classification.

\subsection{Input Representation and Preprocessing}
The input DNA sequence is composed of characters from the alphabet $\{A, C, G, T, N\}$. In the experiments, we map the nucleotides to discrete indices
\[
\mathrm{A}/\mathrm{C}/\mathrm{G}/\mathrm{T}/\mathrm{N} \rightarrow 0/1/2/3/4,
\]
and additionally introduce a padding token indexed as 5. To facilitate batch training and positional alignment, all sequences are standardized to a fixed length of 500: sequences longer than 500 are truncated, whereas shorter sequences are padded. On this basis, the discrete tokens are mapped into dense vector representations through an embedding layer, passed through the front-end convolutional module to extract local motif features, and then combined with learnable positional encodings before being fed into the Transformer encoder.

\subsection{Training Configuration}
Model training follows the supervised binary-classification setting, with a sigmoid output greater than 0.5 taken as the threshold for the positive class. We use the Adam optimizer with an initial learning rate of $1\times 10^{-4}$, train for 10 epochs, and set the batch size to 64. A cosine annealing schedule is used for the learning rate, with a minimum learning rate of $1\times 10^{-6}$. To improve training stability, gradient clipping is applied to all parameters before each update, with a maximum norm of 1.0. Unless otherwise noted, the default hyperparameters are as follows: hidden dimension $d_{\mathrm{model}}=128$, number of encoder layers 3, feed-forward dimension 512, dropout 0.1, number of latent hidden units 16, and initial coefficient of latent interaction strength 0.5.

\subsection{Progressive Scheduling of Structural Constraints}
To prevent structural constraints from interfering too early with the classification objective, we adopt a progressive scheduling strategy for both discrete gate sampling and energy regularization. Specifically, the gating variables are approximately discretized via Gumbel-Softmax, whose temperature parameter $\tau$ is gradually annealed from 1.0 to 0.5. Soft sampling is used during the first three epochs, after which hard sampling is enabled. Meanwhile, the weight of the energy term is fixed at 0 during the first three epochs and then increases linearly from the fourth epoch onward until it reaches the preset value of 0.1. This warm-up design allows the model to first learn a stable classification boundary and then gradually incorporate interaction-structure constraints, yielding a smoother optimization process that balances classification performance and structural interpretability.

\subsection{Baselines}
Because the currently completed and reproducible experimental results are concentrated on the comparison of the main models under a unified training setup, the final paper retains only the following three categories of baselines, around which the subsequent analysis is organized:
\begin{enumerate}[label=(\arabic*)]
    \item Provider CNN baseline: the public PyTorch CNN result for \texttt{human\_enhancers\_cohn} provided by the Genomic Benchmarks project is used as an external baseline \cite{gresova2023genomic,genomicbenchmarks_cnn_notebook};
    \item Plain Transformer: the input embedding, convolutional front end, number of encoder layers, and training hyperparameters are kept unchanged, and only the structured gated attention is replaced with standard \texttt{MultiheadAttention};
    \item Full BM-Transformer: the complete Boltzmann-machine-enhanced Transformer proposed in this paper.
\end{enumerate}
These three model groups are adopted because they correspond, respectively, to the classical convolutional baseline reported by the public dataset provider, the standard Transformer baseline without a structural prior, and the structured-attention model proposed here. Together, they allow us to answer two central questions more clearly: whether the Transformer outperforms the traditional convolutional baseline, and whether the Boltzmann-style structural prior provides additional modeling value while maintaining performance.

\subsection{Evaluation Metrics}
Given the completeness of the available experimental records, we use Accuracy as the primary evaluation metric and additionally report Final Train Acc, Final Val Acc, Best Val Acc or public Test Acc, and Best Val Loss. Because the Provider CNN baseline comes from a public notebook, only the reported test accuracy and corresponding loss are directly available; accordingly, entries that do not have a counterpart are marked as ``--''. This presentation allows the public CNN baseline and the self-trained Transformer-family models to be included in a single main results table without conflating different evaluation protocols.

\section{Results and Analysis}
\subsection{Main Results}
\begin{figure}[!htbp]
\centering
\begin{minipage}[t]{0.48\textwidth}
    \centering
    \safeincludegraphics[width=\linewidth]{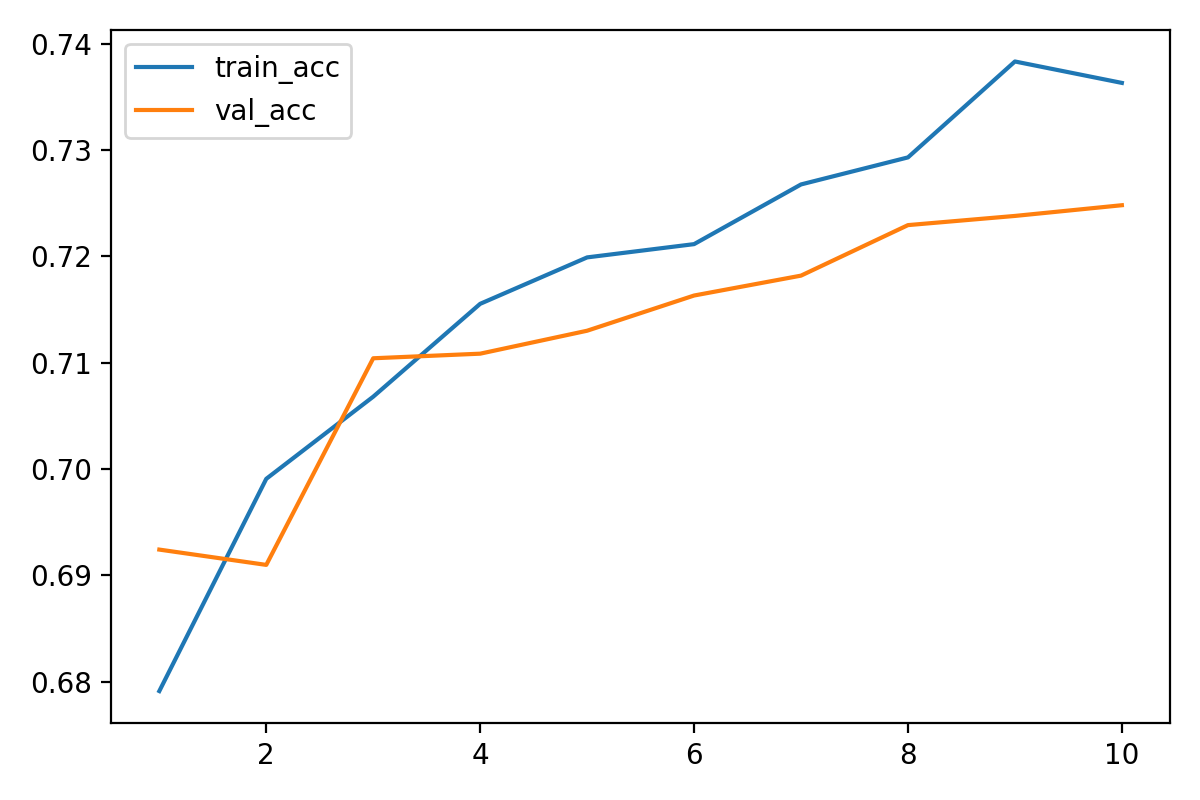}
\end{minipage}\hfill

\begin{minipage}[t]{0.48\textwidth}
    \centering
    \safeincludegraphics[width=\linewidth]{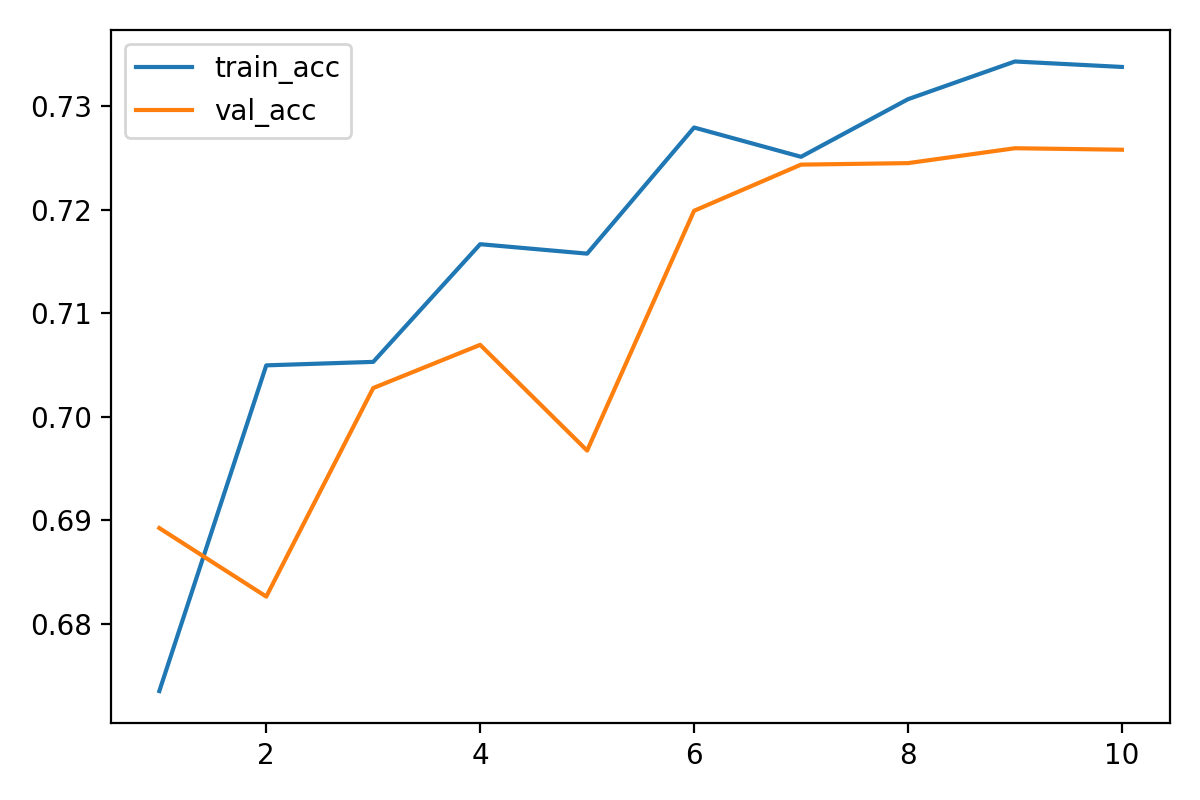}
\end{minipage}
\caption{Comparison of training accuracy curves. The left panel shows the Full BM-Transformer and the right panel shows the Plain Transformer. Both models converge stably and achieve similar validation accuracies near epoch 10.}
\label{fig:acc_compare}
\end{figure}

Table~\ref{tab:main_compare} and Figure~\ref{fig:acc_compare} summarize the main results of the three model classes on the \texttt{human\_enhancers\_cohn} dataset. First, compared with the public CNN baseline reported by the dataset provider, both Transformer architectures achieve higher accuracy: the public test accuracy of the Provider CNN is 0.6950, whereas the best validation accuracies of the Plain Transformer and Full BM-Transformer reach 0.7260 and 0.7248, respectively. Second, between the two Transformer variants, the Plain Transformer is slightly higher than the Full BM-Transformer in both final validation accuracy and best validation accuracy, although the difference is very small. Meanwhile, the Full BM-Transformer attains a slightly higher final training accuracy than the Plain Transformer (0.7363 vs.\ 0.7338). Taken together, these results support two relatively robust conclusions: first, Transformer-family models substantially outperform the public CNN baseline on this task; second, the proposed BM-Transformer performs essentially on par with the standard Transformer in classification accuracy, and its primary value is better described as introducing explicit structural modeling while maintaining competitive accuracy, rather than as achieving a dramatically higher classification metric.

\begin{table}[!htbp]
\centering
\caption{Comparison of the main results. For the self-trained models in this paper, ``Best Val Acc / Test Acc'' denotes the best validation accuracy; for the Provider CNN, this column reports the test accuracy from the public notebook.}
\label{tab:main_compare}
\begin{tabular}{lcccc}
\toprule
Model & Final Train Acc & Final Val Acc & Best Val Acc / Test Acc & Best Val Loss \\
\midrule
Provider CNN (PyTorch) & 0.7160 & -- & 0.6950 & 0.6267 \\
Plain Transformer & 0.7338 & 0.7258 & 0.7260 & 0.5371 \\
Full BM-Transformer & 0.7363 & 0.7248 & 0.7248 & 0.5397 \\
\bottomrule
\end{tabular}
\end{table}

\subsection{Comparison with the Public CNN Baseline}
The Provider CNN baseline represents the reproducible result of a local convolutional model from the public repository associated with this benchmark dataset. Relative to this baseline, both the Plain Transformer and the Full BM-Transformer improve the main metric by approximately three absolute percentage points. This result indicates that, on the \texttt{human\_enhancers\_cohn} task, sequence modeling based only on local convolutional receptive fields can already achieve reasonable performance, but the addition of self-attention still yields a stable benefit by enabling the model to exploit longer-range context and cross-site dependencies. Accordingly, the subsequent discussion focuses primarily on the comparison between the Plain Transformer and the Full BM-Transformer, rather than reiterating the general superiority of Transformers over a simple CNN.

\subsection{Comparison Between BM-Transformer and Plain Transformer}
As shown in the main results table, the Full BM-Transformer and the Plain Transformer perform very similarly: the best validation accuracy of the Plain Transformer is 0.7260, whereas that of the Full BM-Transformer is 0.7248, a difference of only 0.0012. At the same time, the Full BM-Transformer attains a slightly higher final training accuracy, but this does not translate into a correspondingly larger improvement on the validation set. This suggests that, under the current data scale and number of training epochs, the Boltzmann-style structural prior has not yet been converted into a clear advantage in classification accuracy. A more appropriate interpretation is that the structural prior alters the intermediate representation of attention, biasing the model toward learning sparse, interpretable, and coupled connectivity graphs, and that on the present task the principal benefit of this inductive bias lies more in structural representation than in a marked increase in final accuracy.

\subsection{Qualitative Observations of the Structured Model}
\begin{figure}[!htbp]
\centering
\safeincludegraphics[width=0.72\textwidth]{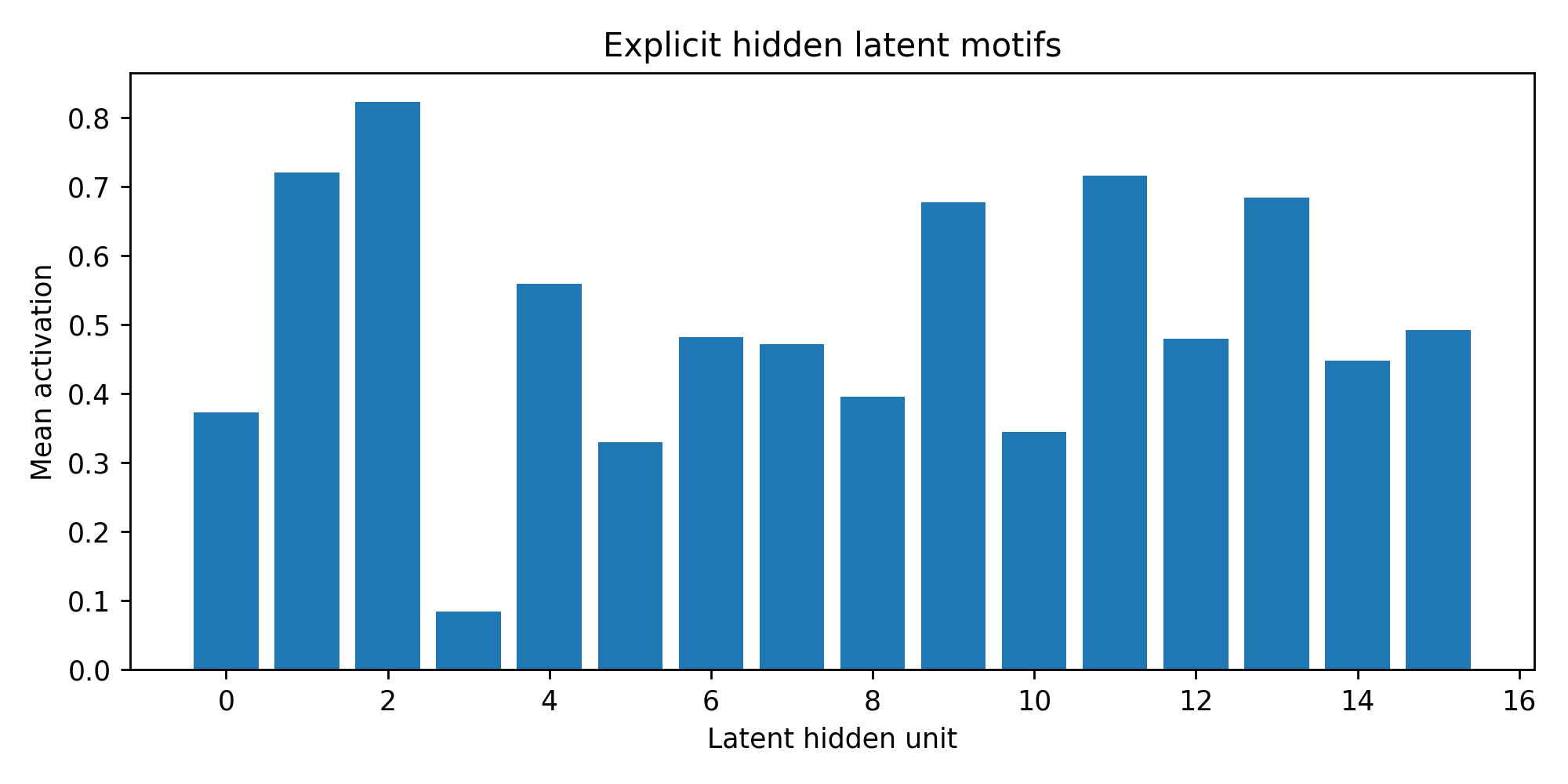}
\caption{Average activation strengths of the 16 latent hidden units. The clear differences across units indicate that the model does not rely uniformly on all latent modules, but instead emphasizes a smaller subset of more active structural components.}
\label{fig:usage_summary}
\end{figure}

\begin{figure}[!htbp]
\centering
\begin{minipage}[t]{0.48\textwidth}
    \centering
    \safeincludegraphics[width=\linewidth]{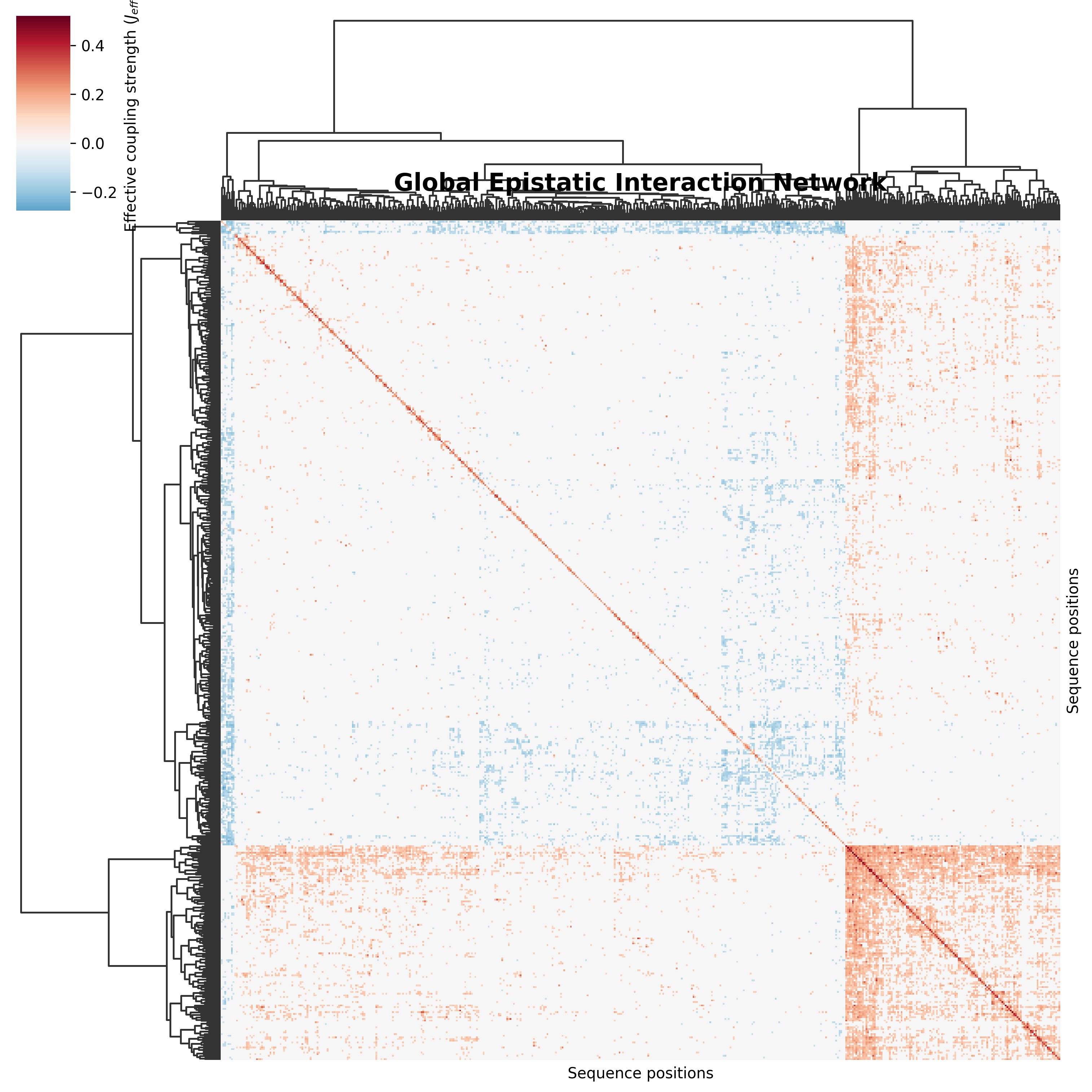}
\end{minipage}\hfill
\begin{minipage}[t]{0.48\textwidth}
    \centering
    \safeincludegraphics[width=\linewidth]{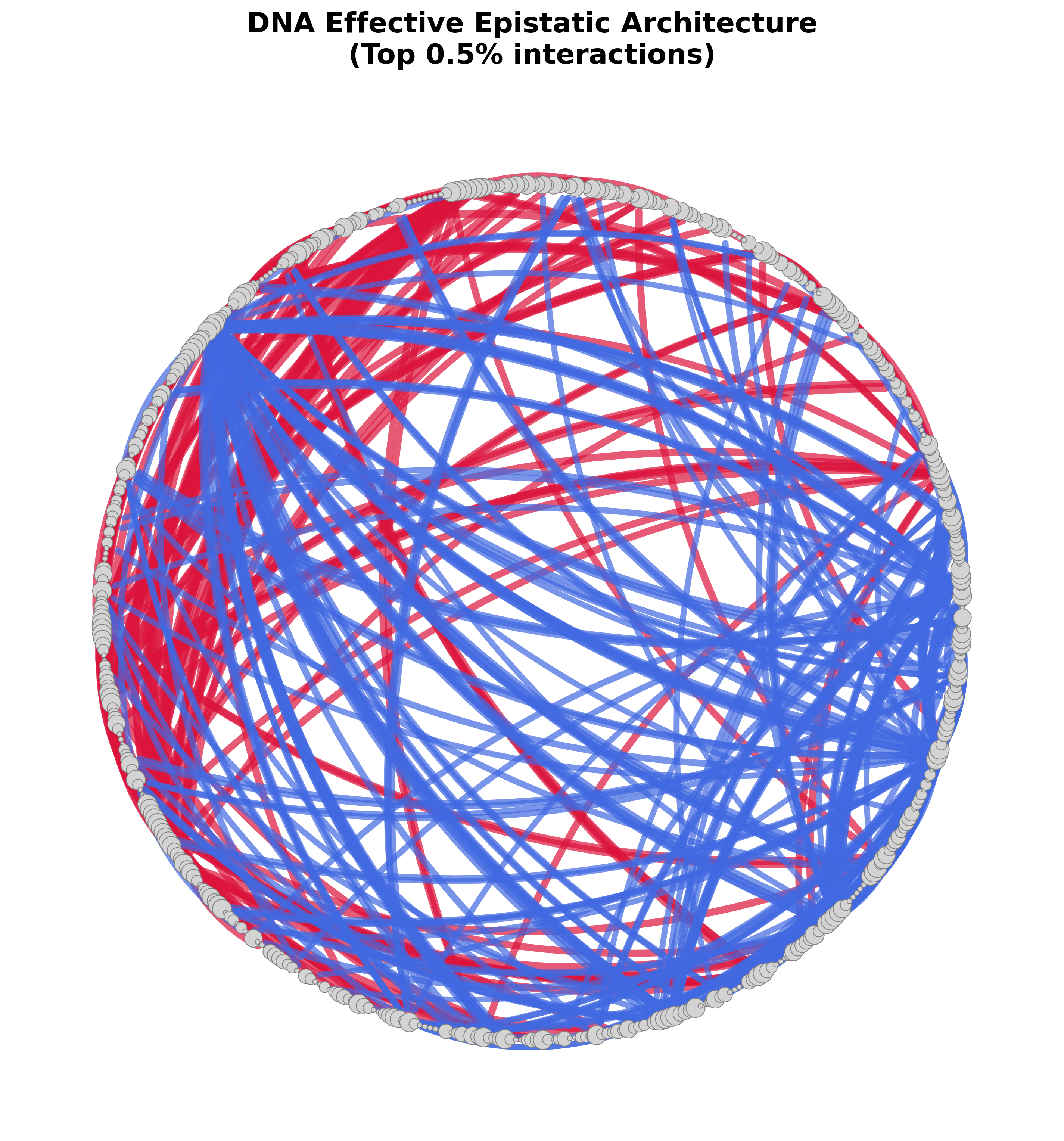}
\end{minipage}
\caption{Visualization of pairwise interactions. The left panel is a clustered heatmap of the effective interaction-strength matrix, showing block-like coupling structures among positions; the right panel is a network formed by the strongest 0.5\% positive and negative interaction edges, with red and blue denoting couplings of different signs.}
\label{fig:pairwise_structure}
\end{figure}

\begin{figure}[!htbp]
\centering
\safeincludegraphics[width=0.94\textwidth]{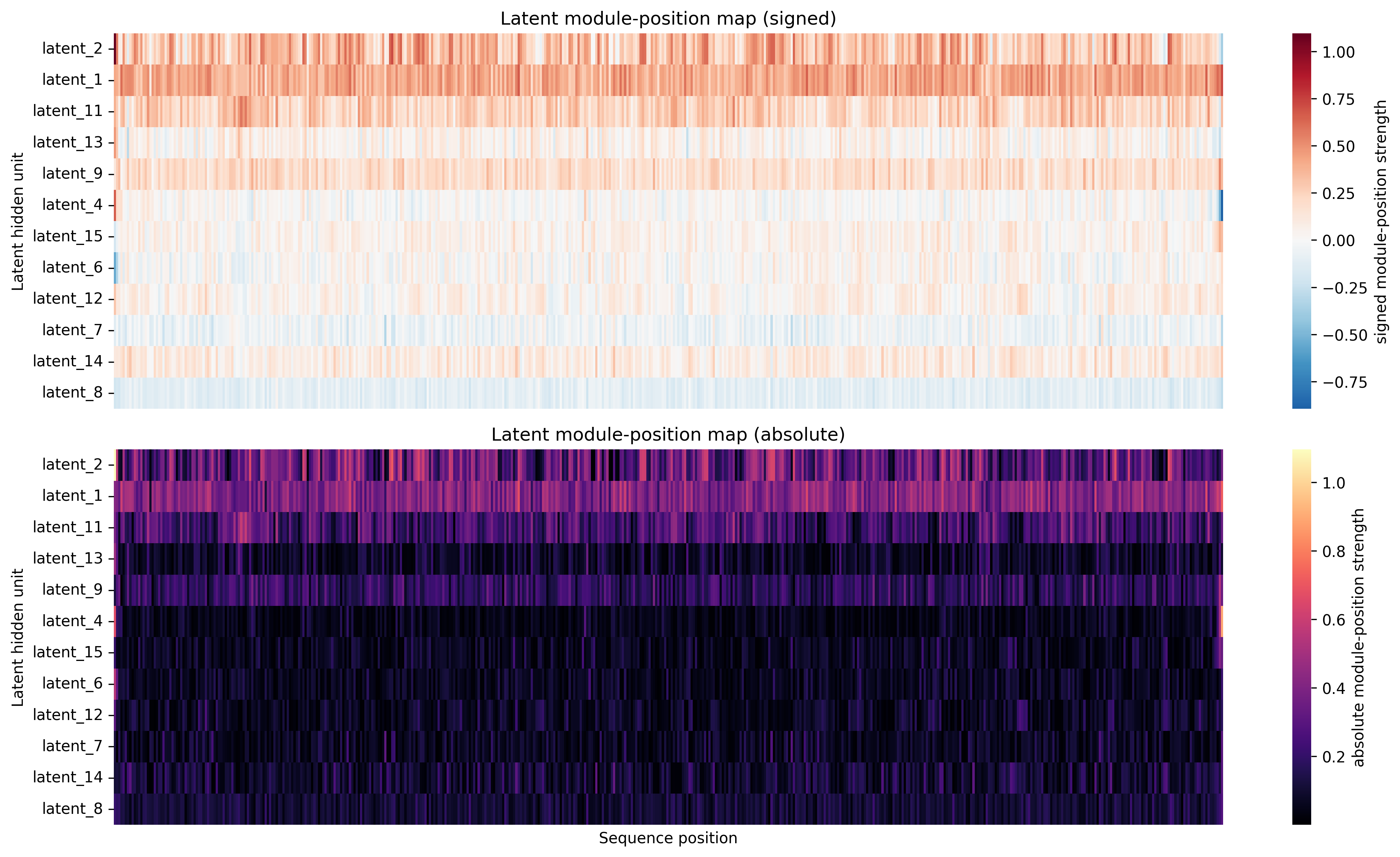}
\caption{Latent module--position map. The upper panel shows the signed module--position interaction strengths, and the lower panel shows their absolute values, which can be used to examine the hotspot distributions of different latent hidden units across the sequence.}
\label{fig:latent_position_map}
\end{figure}

\begin{figure}[!htbp]
\centering
\safeincludegraphics[width=0.92\textwidth]{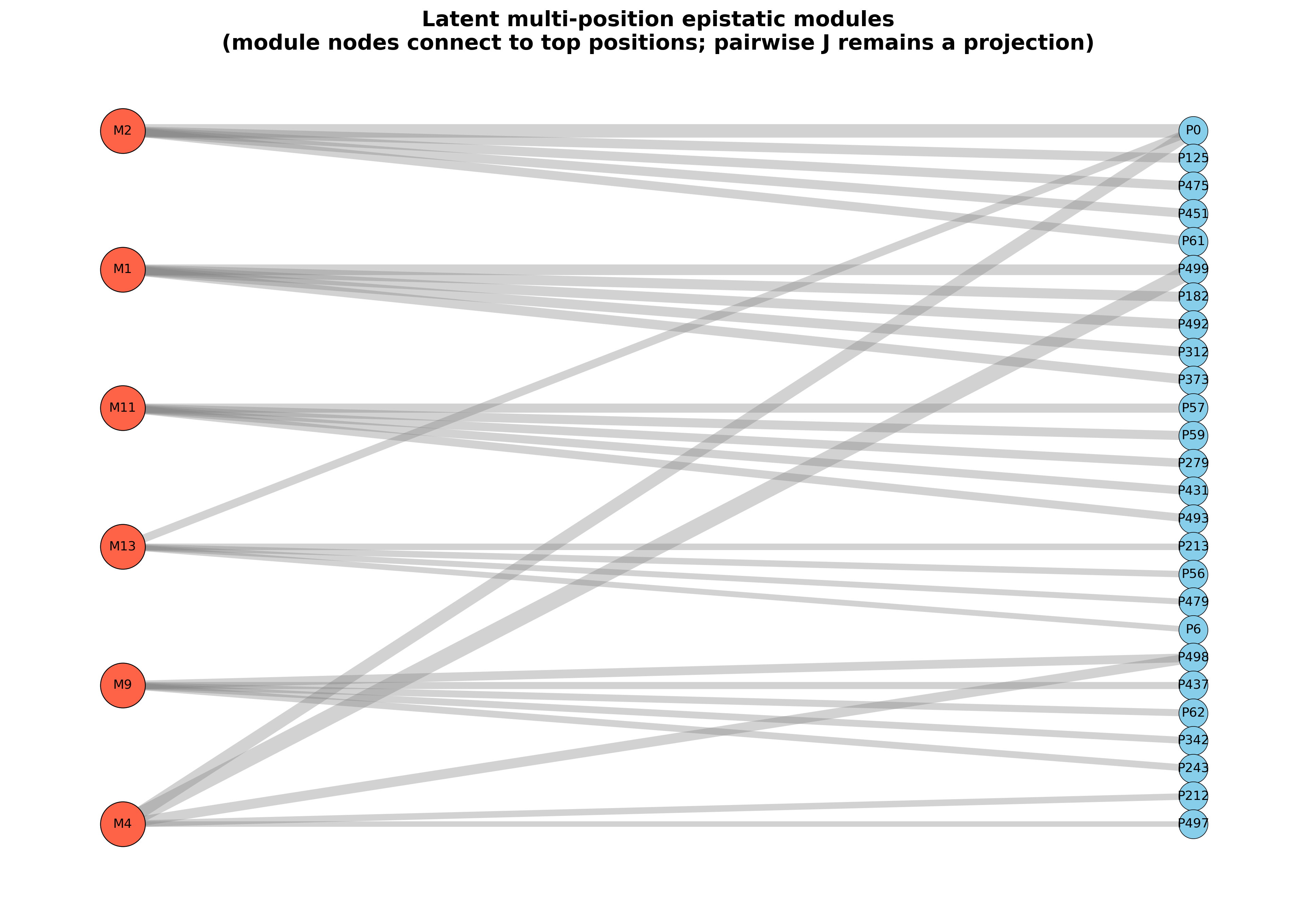}
\caption{Hypergraph visualization of latent higher-order interaction modules. The nodes on the left represent latent modules, and the nodes on the right represent important sequence positions connected with high weights; the fact that one module can connect multiple positions indicates that the model can capture multi-site combinatorial regulatory patterns beyond pairwise projection.}
\label{fig:latent_hypergraph}
\end{figure}

Although the BM-Transformer does not clearly surpass the Plain Transformer in accuracy, visualizations of its internal variables show that it indeed learns relatively clear structural patterns. Figure~\ref{fig:usage_summary} shows substantial differences in the average activation strengths of different latent hidden units, indicating that the model does not use all latent modules uniformly but instead tends to rely on a smaller set of more active structural components. This phenomenon is consistent with the modeling hypothesis that higher-order combinatorial dependencies are dominated by a limited number of modules. Furthermore, the pairwise interaction heatmap and network visualization in Figure~\ref{fig:pairwise_structure} show that the learned effective interactions include both positive and negative couplings and exhibit both local block structure and long-range cross-position edges. Finally, Figures~\ref{fig:latent_position_map} and~\ref{fig:latent_hypergraph} illustrate the correspondence between latent modules and sequence positions, suggesting that the BM-Transformer can indeed organize several important sites into shared latent modules. It should be emphasized that these figures provide qualitative evidence about structured intermediate representations rather than direct evidence of improved classification performance; the appropriate conclusion is that the proposed model learns richer structural objects, not that it outperforms the standard Transformer on every metric.

\section{Discussion}
\subsection{When Structured Attention Is Valuable}
The proposed method is particularly valuable in the following situations:
\begin{enumerate}[label=(\arabic*)]
    \item when the goal is not only prediction but also structural discovery;
    \item when one aims to recover latent interaction graphs among sites;
    \item when data are limited and stronger inductive bias is needed;
    \item when the task exhibits pronounced higher-order combinatorial dependence.
\end{enumerate}

\subsection{Limitations}
\begin{enumerate}[label=(\arabic*)]
    \item mean-field inference is only an approximation and may deviate from the true posterior;
    \item Gumbel-Softmax uses gradients that are biased, although low variance;
    \item the design of the energy function can substantially affect structural quality;
    \item compared with a plain Transformer, training and hyperparameter tuning are more costly;
    \item structural interpretability does not automatically imply biological causality and still requires downstream experimental validation.
\end{enumerate}

\section{Conclusion}
This paper proposes a Boltzmann-machine-enhanced Transformer for DNA sequence classification, whose central idea is to replace standard softmax attention with a Boltzmann-style structural distribution. By introducing local bias terms, pairwise interactions, and latent hidden units, the model extends attention from a mechanism of continuous weight allocation to a process of structured gated-graph inference. Furthermore, we use mean-field variational inference to obtain a continuous approximation to the gating posterior and combine it with Gumbel-Softmax to gradually compress the soft structure into a near-discrete gating pattern, thereby improving structural sharpness while preserving trainability. By jointly optimizing classification loss and energy loss, the model aims not only for predictive accuracy but also for more plausible, lower-energy, and more interpretable structures. Experimental results show that the proposed method remains close to the standard Transformer baseline in classification performance, while its more distinctive advantage lies in providing explicit site-interaction graphs, latent-module responses, and visualizations of higher-order combinatorial structure.

From a theoretical perspective, this paper provides a complete derivation from the energy function to mean-field fixed-point updates, and further to Gumbel-Softmax and the joint loss, showing that the method is grounded in a unified framework of probabilistic graphical modeling and differentiable discrete optimization. Future work may explore stronger inference algorithms, incorporate external biological priors, and extend the method to larger-scale genomic pretraining and causal analysis settings.

\appendix
\section{Complete Summary of Formulas}
\subsection{Standard Softmax Attention}
\begin{align}
A_{hts} &= \frac{\exp\!\left(\mathbf{q}_{ht}^{\top}\mathbf{k}_{hs}/\sqrt{d_h}\right)}{\sum_{s'} \exp\!\left(\mathbf{q}_{ht}^{\top}\mathbf{k}_{hs'}/\sqrt{d_h}\right)}, \\
\mathbf{o}_{ht} &= \sum_s A_{hts}\,\mathbf{v}_{hs}.
\end{align}

\subsection{Boltzmann-Style Structural Distribution}
\begin{align}
p(z\mid x) &= \frac{1}{Z(x)} \exp\!\bigl(-E(z;x)\bigr), \\
Z(x) &= \sum_z \exp\!\bigl(-E(z;x)\bigr).
\end{align}

\subsection{Overall Energy Function}
\begin{align}
E(z,u;x) &= E_{\mathrm{bias}}(z;x) + E_{\mathrm{pair}}(z;x) + E_{\mathrm{latent}}(z,u;x), \\
E_{\mathrm{bias}}(z;x) &= -\sum_h\sum_t\sum_s h_{hts} z_{hts}, \\
E_{\mathrm{pair}}(z;x) &= -\frac{1}{2}\sum_h\sum_t\sum_{s\neq s'} J_{ss'}^{(h,t)} z_{hts} z_{hts'}, \\
E_{\mathrm{latent}}(z,u;x) &= -\sum_h\sum_t\sum_m b_{htm}^{(u)} u_{htm} - \sum_h\sum_t\sum_s\sum_m W_{sm}^{(h,t)} z_{hts} u_{htm}.
\end{align}

\subsection{Mean-Field Approximation}
\begin{align}
q(z,u) &= \prod_h\prod_t\prod_s q_{hts}(z_{hts}) \cdot \prod_h\prod_t\prod_m r_{htm}(u_{htm}), \\
q_{hts}(z_{hts}=1) &= s_{hts}, \\
r_{htm}(u_{htm}=1) &= r_{htm}.
\end{align}

\subsection{Variational Free Energy}
\begin{equation}
\mathcal{F}(q)=\mathbb{E}_q\bigl[E(z,u;x)\bigr]-H(q).
\end{equation}

\subsection{Fixed-Point Updates}
\begin{align}
s_{hts} &= \sigma\!\left(h_{hts} + \sum_{s'\neq s} J_{ss'}^{(h,t)} s_{hts'} + \sum_m W_{sm}^{(h,t)} r_{htm}\right), \\
r_{htm} &= \sigma\!\left(b_{htm}^{(u)} + \sum_s W_{sm}^{(h,t)} s_{hts}\right).
\end{align}

\subsection{Boltzmann-Machine-Gated Attention Output}
\begin{equation}
\mathbf{o}_{ht} = \frac{\sum_s s_{hts}\,\mathbf{v}_{hs}}{\sum_s s_{hts}+\varepsilon}.
\end{equation}

\subsection{Gumbel-Softmax}
\begin{align}
\alpha_{hts}^{(0)} &= \log(1-s_{hts}), \\
\alpha_{hts}^{(1)} &= \log(s_{hts}), \\
\tilde{z}_{hts}^{(c)} &= \frac{\exp\!\left((\alpha_{hts}^{(c)} + g_{hts}^{(c)})/\tau\right)}{\sum_{c'\in\{0,1\}} \exp\!\left((\alpha_{hts}^{(c')} + g_{hts}^{(c')})/\tau\right)}.
\end{align}

\subsection{Task Loss}
\begin{equation}
\mathcal{L}_{\mathrm{task}} = -\Bigl[y\log\sigma(\hat{y}) + (1-y)\log\bigl(1-\sigma(\hat{y})\bigr)\Bigr].
\end{equation}

\subsection{Energy Loss}
\begin{equation}
\mathcal{L}_{\mathrm{energy}} = \max\bigl(0, E(s_{\mathrm{pos}}) - E(s_{\mathrm{neg}}) + m\bigr).
\end{equation}

\subsection{Overall Loss}
\begin{equation}
\mathcal{L} = \mathcal{L}_{\mathrm{task}} + \lambda \, \mathcal{L}_{\mathrm{energy}}.
\end{equation}

\end{document}